\pdfoutput=1

\documentclass[11pt]{article}

\usepackage{ACL2023}

\usepackage{times}
\usepackage{latexsym}
\usepackage{makecell}
\usepackage{booktabs}
\usepackage{multirow}
\usepackage{float}
\usepackage{ltablex}
\usepackage{array}
\usepackage{graphicx}
\usepackage{soul}
\usepackage{amsmath}
\usepackage{subcaption}
\DeclareMathOperator*{\argmax}{argmax}
\usepackage{makecell}
 \usepackage{xcolor}

\usepackage{cleveref}
\crefformat{section}{\S#2#1#3} 
\crefformat{subsection}{\S#2#1#3}
\crefformat{subsubsection}{\S#2#1#3}

\usepackage[T1]{fontenc}

\usepackage[utf8]{inputenc}

\usepackage{microtype}

\usepackage{inconsolata}

\renewcommand{\footnotesize}{\fontsize{8pt}{11pt}\selectfont}

\usepackage{bbding}
\usepackage{pifont}
\usepackage{wasysym}
\usepackage{amssymb}
\newcommand{\cmark}{\ding{51}}%
\newcommand{\xmark}{\ding{55}}%

\usepackage[most]{tcolorbox}
\colorlet{Blue}{blue!40!}
\definecolor{RedClear}{rgb}{0.96,0.68,0.81}
\tcbset{on line, 
        boxsep=2pt, left=0pt,right=0pt,top=0pt,bottom=0pt,
        colframe=white,colback=RedClear        
        }

\title{Semantic Similarity Models for Depression Severity Estimation}

\author{Anxo Pérez $^{1}$, Neha Warikoo$^{2}$, Kexin Wang$^{2}$, Javier Parapar$^{1}$, Iryna Gurevych$^{2}$ \\
$^1$Information Retrieval Lab \\
  Centro de Investigación en Tecnoloxías da Información e as Comunicacións (CITIC) \\
  Universidade da Coruña, A Coruña, Spain \\
  \url{https://irlab.org/} \\
 $^2$Ubiquitous Knowledge Processing Lab (UKP Lab) \\
  Department of Computer Science and Hessian Center for AI (hessian.AI) \\
  Technical University of Darmstadt \\
  \url{www.ukp.tu-darmstadt.de}
}

\begin{document}
\maketitle
\begin{abstract}

Depressive disorders constitute a severe public health issue worldwide. However, public health systems have limited capacity for case detection and diagnosis. In this regard, the widespread use of social media has opened up a way to access public information on a large scale. Computational methods can serve as support tools for rapid screening by exploiting this user-generated social media content. This paper presents an efficient semantic pipeline to study depression severity in individuals based on their social media writings. We select test user sentences for producing semantic rankings over an index of representative training sentences corresponding to depressive symptoms and severity levels. Then, we use the sentences from those results as evidence for predicting symptoms severity. For that, we explore different aggregation methods to answer one of four Beck Depression Inventory (BDI-II) options per symptom. We evaluate our methods on two Reddit-based benchmarks, achieving 30\% improvement over state of the art in terms of measuring depression level\footnote{Implementation at \textit{(restricted for anonymity)}.}.

\end{abstract}

\section{Introduction}
\label{sec:introduction}

Around two-thirds of all cases of depression remain undiagnosed according to conservative estimates~\cite{epstein2010didn}. To help with this problem, governments and agencies have launched programs to raise awareness of mental health in their citizens~\cite{arango2018preventive}. In this context, detecting and receiving appropriate treatment in the early stages of these diseases is essential to reduce their impact and case escalation~\cite{picardi2016randomised}. However, the insufficient resources of the public health systems severely limit their capacity for case detection and diagnosis.

As an alternative to public health systems, social platforms are a promising channel to assess risks in an unobtrusive manner~\cite{choudhury-et-al-2013}, where people tend to consider these platforms as comfortable media to express their feelings and concerns~\cite{chancellor2020methods}. Exploiting this type of user-generated content, NLP techniques have shown promising results in terms of identifying depressive patterns and linguistic markers~\cite{rissola-2021-survey}. Due to the growing popularity of the mental health detection models, the community also produced diverse datasets~\cite{yates-etal-2017-depression, cohan-etal-2018-smhd}, with the Early Risk Prediction on the Internet (eRisk)~\cite{eriskbook}, and the Computational Linguistics and Clinical Psychology (CLPsych)~\cite{clpsych-2022-linguistics} being the two most popular benchmarks in the field. They produce task definitions, datasets and evaluation methodologies to encourage research in this domain.

Depression identification from social media posts faces challenges considering their integration into clinical settings~\cite{10.1093-jamiaopen-ooz054}. Previous studies formulated this task as a binary classification problem (i.e., depressed vs control users)~\cite{rissola-2021-survey}. Despite achieving remarkable results under this setting, ignoring different levels of depression limits the capacity to prioritize users with higher risks~\cite{naseem2022early}. Moreover, most existing approaches focused on the use of engineered features, which may be more difficult to interpret than other clinical markers\footnote{An observable sign indicative of a depressive tendency.}, such as the integration of recognized depressive symptoms~\cite{mowery2017understanding}. Similarly, the black-box nature of deep learning models also limits the ability to understand their decisions, especially by domain experts like clinicians.

In this paper, we perform a fine-grained analysis of depression severity using semantic features to detect the presence of symptom markers. Our methods adhere to accepted clinical protocols by automatically completing the BDI-II~\cite{dozois1998psychometric}, a questionnaire used to measure depression. The BDI-II includes 21 recognized symptoms, such as sadness, fatigue or sleep issues. Each symptom has four alternative responses scaled in severity from 0 to 3. Using a sentence-based pipeline, we build 21 different symptom-classifiers for estimating the user responses to the symptoms. For this purpose, we employ eRisk collections related to depression levels~\cite{losada2019overview,losada2020erisk,parapar2021overview}. In our pipeline, we explore selection algorithms to filter relevant sentences to each BDI-II symptom from training users. Once filtered, we index these training sentences with the user responses as labels (0 - 3) as examples of how people with different severity speak about the symptom. Then, to predict test users responses, we select their relevant sentences, which serve as queries to produce a semantic ranking over the indexed training sentences. Finally, we construct two aggregation methods based on the ranking results to estimate the symptoms severity.


The main contributions of this work are: $1$) We present a semantic retrieval pipeline to perform a fine-grained classification of the severity of depressive symptoms. Following the symptoms covered by the BDI-II, our methods also consider different depression severity levels, distinguishing between lower and higher risks. $2$) We propose a data selection process using a range of unsupervised and semi-supervised selection strategies to filter relevant sentences for the symptoms. 
$3$) Experiments using different variants of our pipeline achieved remarkable results in two eRisk collections, outperforming state of the art considering depression severity in both datasets.

\section{Related Work}
\label{sec:related_work}

Extensive research investigated the use of engineered features to identify linguistic markers and patterns related to mental disorders from different social platforms~\cite{gaur2021characterization,coppersmith-etal-2014-quantifying,yates-etal-2017-depression}. For instance, the LIWC writing analysis tool~\cite{pennebaker2003psychological}, equipped with psychological categories, revealed remarkable differences in writing style between depression and control groups~\cite{choudhury-et-al-2013}. Other studies used depression and emotional lexicons to determine depression markers~\cite{cacheda2019early}, whereas \citet{trotzek-2018} examined additional distinctive features, leveraging profile metadata (e.g. posting hours or posts length) and social activity to examine the mental state of individuals.


The recent advances in contextualized embeddings significantly impacted many NLP-related tasks, including depression detection in social media. These deep learning models have consistently outperformed engineered features on diverse datasets~\cite{jiang-etal-2020-detection,Nguyen-et-al-2022-leveraging-symptoms}. However, they lack the interpretability clinicians require to rely on the results from automated screening methods~\cite{Amini2020TowardsEI}. To enhance interpretability, the works proposed to the \textit{eRisk depression estimation shared task} based their efforts on predicting BDI-II symptoms responses~\cite{uban2020deep, spartalis-transfer-2021,UPV-symanto-2021}. This study is directly related to eRisk, as we work with the eRisk collections and follow the same evaluation methodology. In contrast to these approaches, our methods highlight the user posts that lead to every symptom decision, which may be helpful for further inspection of model predictions.

Besides the works presented to eRisk, two recent studies explored the use of depressive symptoms to screen social media posts. \citet{Zhang-2022-Depression-Scale} aggregated symptoms from different questionnaires into a BERT-base model to calculate symptom risk at post level. \citet{Nguyen-et-al-2022-leveraging-symptoms} experimented with various methods using symptom markers to detect depression, demonstrating their potential to improve the generalization and interpretability of their approaches. In this case, authors considered the symptoms from the PH9Q questionnaire \cite{kroenke2001phq} to define manual pattern-based strategies and train symptom-classifiers at post level. Both approaches formulated their methods with a binary classification setting, while our approach considers different severity levels. We also differ in that we pre-compute dense representations of training posts, rather than relying on pre-trained language models, which may be slow for many practical cases \cite{reimers-2019-sentence-bert}. Our approach only needs a few post encodings and cosine similarity calculations, improving the efficiency of our solutions.

\begin{figure*}[h]
\centering
\includegraphics[width=0.9\textwidth]{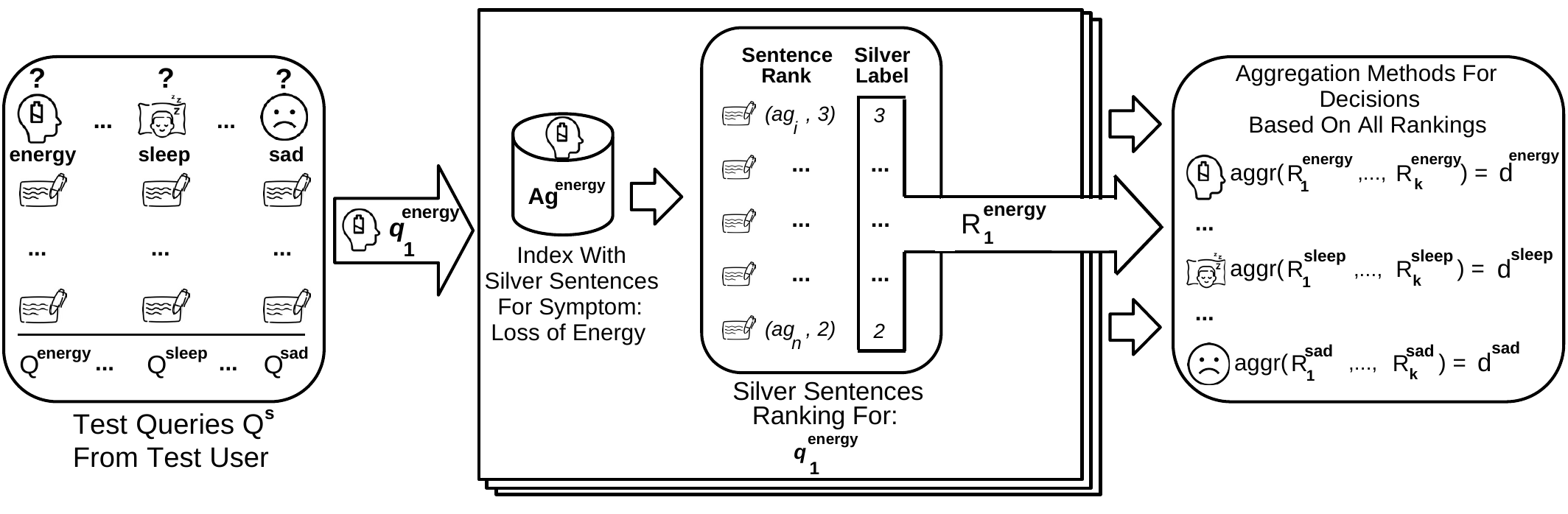}
\caption{Retrieval pipeline to predict symptom options for a test user. $R_{1}^{energy}$ is the list with the top ranked silver sentences for the query $q_{1}^{energy}$. Each silver sentence from the rank has a silver label associated (0-3). $d^{energy}$ represents the option decision for that symptom based on the ranking retrieved for all the test queries, $Q^{energy}$.}
\label{fig:workflow_pipeline}
\end{figure*}

\section{Method}
\label{sec:proposal}

\textbf{Problem definition: } We aim to estimate the depression severity level for users based on the Writing History (WH) of their social media posts. We define depression severity levels following the clinical classification schema of the BDI-II score~\cite{lasa2000use}. The score is the sum of the option responses to the 21 symptoms covered by this questionnaire, and it is associated with four depression levels. Table~\ref{tab:bdi_levels} shows these levels.

\begin{table}[t]
        \centering
        \small
        \begin{tabular}{lc}
        \toprule
        \textbf{Depression level} & \textbf{BDI-II Score} \\ 
        \midrule 
         Minimal depression & (0-9)    \\
         Mild depression  &  (10-18)   \\
         Moderate depression  & (19-29)   \\
         Severe depression  & (30-63)    \\
        \bottomrule
        \end{tabular}
        \caption{Depressive levels related to the BDI-II score.}
        \label{tab:bdi_levels}
        \vspace{-2mm} 
\end{table}

Instead of relying on a unique classifier to calculate that score, we build 21 different symptom-classifiers (i.e., one for each symptom). For this purpose, we categorize the symptoms into one of its four response options. Therefore, we formulate each classifier as a multi-class classification problem. Table~\ref{tab:example_item_sentences} provides an example of the option descriptions for the symptom \textit{Loss of energy}. To estimate the depression level for a user, we aggregate the predicted responses of all the symptom-classifiers.

Our approach relies on two critical components: $1)$ a semantic retrieval pipeline (\cref{subsec:semantic-retrieval-pipeline}) and $2)$ silver sentences selection (\cref{subsec:silver-labels-generation}). The symptom-classifiers follow a semantic retrieval pipeline to predict every symptom decision. This pipeline searches for semantic similarities over an index of silver sentences for a specific symptom $s$, denoted as $Ag^{s}$. These silver sentences are considered relevant to $s$, and each one has a label corresponding to the symptom options\footnote{Throughout the rest of the paper, we will refer to these severity options as the labels of the symptoms.}, $o$. Formally, $Ag^{s}$ is the set containing the pairs of the silver sentences $ag_{i}$ and their corresponding label $o_{i}$ for the symptom $s$, where $Ag^s= \{(ag_{i}, o_{i})\}$, and $o_{i} \in \{0, 1, 2, 3\}$.

\begin{table}[t]
        \centering
        \small
        \begin{tabular}{>{\centering}p{0.1\columnwidth}p{0.8\columnwidth}}
        \toprule
        \textbf{Option} & \textbf{Description} \\ 
        \midrule 
         \textbf{0} & I have as much energy as ever    \\
           \textbf{1}  &  I have less energy than I used to have   \\
          \textbf{2}  & I do not have enough energy to do very much    \\
          \textbf{3}  & I do not have enough energy to do anything    \\
        \bottomrule
        \end{tabular}
        \caption{BDI-II options for the symptom \textit{Loss of energy}.}
        \label{tab:example_item_sentences}
        \vspace{-2mm} 
\end{table}

To the best of our knowledge, there are no datasets in the literature where sentences are relevant to the symptom and labelled by their severity. For this reason, we propose a selection process to create the silver sentences, $Ag^s$, where we use as training data the eRisk collections (\cref{subsec:silver-labels-generation}). In our experiments, we explore the performance of the semantic pipeline with our generated silver sentences. However, we could apply this pipeline to any similar datasets. In the following subsections, we explain both components in detail.

\subsection{Semantic Retrieval Pipeline}
\label{subsec:semantic-retrieval-pipeline}

Using the writing history from a test user as input, our semantic retrieval pipeline classifies its label severity for a specific symptom $s$. From the publications of the test user, we first select the sentences that are relevant to $s$, which will serve as queries. We denote these relevant sentences as the symptom test queries $Q^{s}$, where $Q^{s} =  \{q_{1}^{s}, ..., q_{k}^{s}\}$, since we select a top $k$ of them. In the next subsection, we explain our sentence selection algorithms (\cref{subsec:silver-labels-generation}). The top $k$ of queries are the input to our semantic pipeline. Figure~\ref{fig:workflow_pipeline} illustrates this process, exemplified for one test query, $q_{1}^{energy}$, corresponding to the symptom \textit{Loss of energy}: 



\begin{figure*}[h]
\centering
\includegraphics[width=0.81\textwidth]{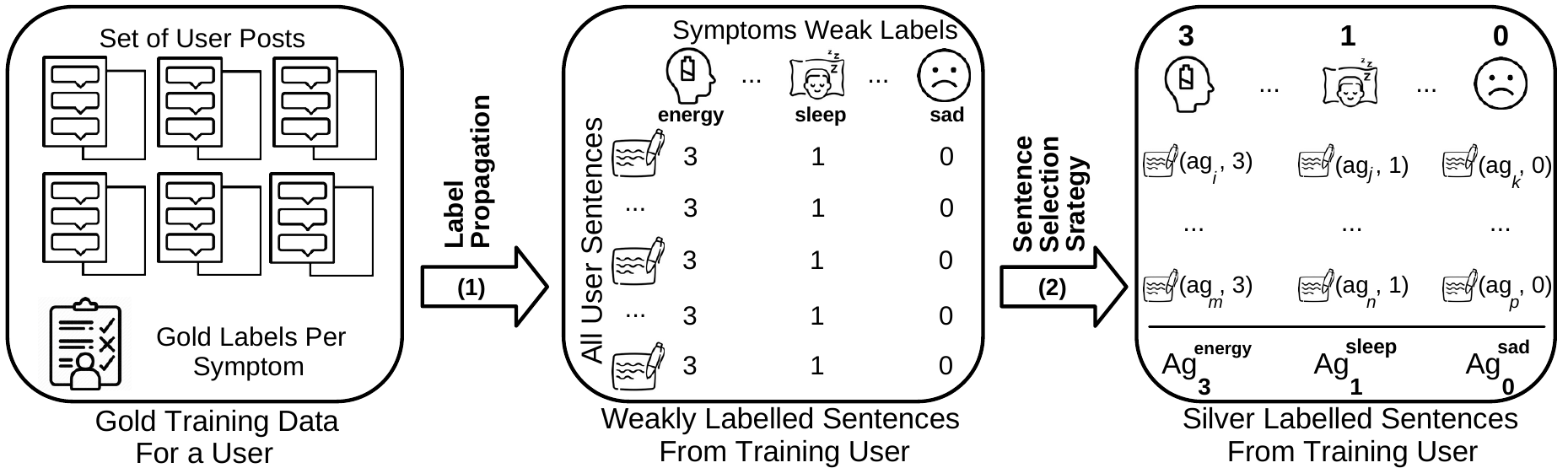}
\caption{From the responses of the eRisk training users to the symptom options (0-3), the silver selection process creates one different set of silver sentences relevant to each symptom $s$ and option $o$, denoted as $Ag_o^s$.}
\label{fig:weak_labels_filter}
\end{figure*}

$1$) The first step consists in calculating a semantic ranking for each test query for the symptom $s$, defined as $q_{i}^{s}$. To calculate that ranking, we encode $q_{i}^{s}$ and all the silver sentences in $Ag^{s}$ for the symptom as embeddings. Then, we use k-Nearest Neighbours (kNN) to compute the semantic similarity of each silver sentence w.r.t the test query $q_{i}^{s}$. The semantic similarity $sm$ for a silver sentence $ag_{j}$, belonging to $Ag^{s}$, and a test query $q_{i}^{s}$, is the cosine similarity between their embeddings ($\phi$): 

\begin{equation}
 sm(ag_{j}, q_{i}^{s}) = cos\big(\phi(ag_{j}),\phi(q_{i}^{s}))    
\end{equation}

Computing $sm$, we produce a ranking of silver sentences, $R_{i}^{s}$, for each test query $q_{i}^{s}$. The silver sentences in the ranking have an associated silver label. For example, the position $j$ of the ranking contains the pair: $R_{i}^{s}[j]= \{(ag_{j}, o_{j})\}$, with $o_{j} \in \{0,1,2,3\}$. To select the cut-off of the rankings $R_{i}^{s}$, we experimented with a varying number of similarity thresholds. To calculate the embeddings, we use a pre-trained model based on RoBERTa\footnote{huggingface.co/sentence-transformers/all-roberta-large-v1} using sentence-transformers (SBERT).


$2$) In the second step, we apply aggregation methods to accumulate the score of the labels based on the ranking results. After processing all the test queries, the final decision predicted for the symptom \textit{s}, $d^{s}$, is the label with the highest accumulated score. We explore two aggregation methods:

\textbf{Accumulative Voting:} For each ranking $R_{i}^{s}$, we count the option labels from the $n$ pairs that are in the rank: $\{(ag_{j}, o_{j})\}$. The label of each silver sentence, $o_{j}$, represents a vote for that option. Then, return the sum of all the votes over the rankings. The final decision for the symptom $s$ is the label with most votes, $d^s = \underset{o}{\argmax}  \, f_{av}(o)$, where:

\begin{align}
\scriptsize
 f_{av}(o) = \sum_{i \in R_{i}^{s} }{\sum_{j=1}^{n}{\begin{cases}
  1 & \{(ag_{j}, o_{j}) |  o_{j} = o\}   \\
  0 & \text{otherwise}
  \end{cases} } }
\end{align}

\textbf{Accumulative Recall:} For each ranking $R_{i}^{s}$, compute the recall for each label $o$. That is, the fraction of silver sentences in the ranking out of all the available silver sentences from that label, denoted as $Ag^s_{o}$, where $Ag^s_{o} = \{(ag_{i}, o_{i}) |  o_{i} = o\}$. Then, we accumulate the recall over the rankings $R_{i}^{s}$. The final decision is $d^s = \underset{o}\argmax f_{ar}(o)$ with:

\begin{align}
\scriptsize
 f_{ar}(o) = \sum_{i \in R_{i}^{s} }{\frac{\sum_{j=1}^{n}{\begin{cases}
  1 &  \{(ag_{j}, o_{j}) |  o_{j} = o\}   \\
  0 & \text{otherwise}
  \end{cases} }}{|Ag^s_{o}|}} 
\end{align}

\subsection{Silver Sentences Selection}
\label{subsec:silver-labels-generation}


We design a process to select relevant sentences for each symptom $s$, and the severity labels $o$ (previously denoted as $Ag^{s}_{o}$), defined as silver sentences. For this purpose, we use the eRisk collections as training data. These collections contain users from the Reddit platform, and have two main elements: $i$) the user responses to the BDI-II symptoms and $ii$) their posts from Reddit. We use the option responses from the users as the severity labels for each symptom ($0-3$). Therefore, the training labels are initially available at user level. Details of eRisk datasets are provided in Section \ref{subsec:erisk-datasets}.

Figure~\ref{fig:weak_labels_filter} illustrates the sentence selection process for one training user and three symptoms. $1)$ In the first step, we propagate the user responses as labels for all the sentences from its writing history, resulting in weakly labelled sentences. For example, in the second component of Figure~\ref{fig:weak_labels_filter}, the user replied with the option $3$ for the symptom \textit{Loss of energy} (first column). Thus, all the sentences from the user have that weak label assigned. However, since users tend to talk about different topics, most of their sentences are not relevant to any symptom. For this reason, the weak labels contain many false positives that introduce noise. $2)$ To reduce this noise,  we propose two distant supervision strategies for sentence selection. These strategies aim to filter out the training sentences that may be non-informative w.r.t the assigned weak label. We implement two different strategies:

\textbf{Option descriptions as queries:} This strategy works in an unsupervised manner, since we consider the option descriptions from the symptoms as queries to select the silver sentences. Table \ref{tab:example_item_sentences} shows an example of the descriptions for the symptom \textit{Loss of energy}. We use each option description as one different query. Based on the sentences retrieved from these queries, we select a top of sentences from the eRisk training users who answered the same option used as the query. Following this approach, we perform lexical and semantic retrieval variants. For lexical search, we use BM25~\cite{robertson1995okapi} to retrieve relevant sentences for each training user. In the semantic variant, we calculate the similarity based on a semantic threshold, as described in the semantic ranking (\cref{subsec:semantic-retrieval-pipeline}), using the same RoBERTa model for the semantic search.

\textbf{Few manually labelled sentences as queries:} A drawback in using the option descriptions of the BDI-II symptoms as queries is that they only have subtle differences among one another. Consequently, previous queries struggle to capture their actual distinctions. To alleviate this problem, we hypothesize that using actual sentences from eRisk training users who answered each option may be better to differentiate between such options. In this second strategy, a small set of manually labelled sentences, referred to as \textit{golden sentences}, serve as queries to generate an augmented silver set. The use of a larger, higher-quality set of queries allows us to cover more diverse expressions of symptom signals.

We used the eRisk2019 training users to obtain the golden sentences. Following the approach by \citet{karisani-2018-health-mentions}, three experts in the field conducted the annotation process. The number of golden sentences was low, averaging 35 per symptom. The data augmentation process consisted of, for every golden sentence belonging to a specific option, following the semantic ranking (\cref{subsec:semantic-retrieval-pipeline}) over the rest of the weakly-labelled sentences from that same option. The final set of relevant sentences combines the golden and the silver sentences that surpass the similarity threshold. Table~\ref{table:augmented-sentences-example} shows an example of a golden sentence along with the top $3$ augmented silver sentences. The golden sentence corresponds to the option $3$ for the symptom \textit{Pessimism in the future}, and the augmented silver sentences correspond to other training users who reported the same option\footnote{Information about the dataset construction and the annotation process can be found in Appendix \ref{appendix-sec:annotation-process}.}.

\begin{table}[t]
    \centering
    \LARGE
    \resizebox{\linewidth}{!}{%
    \begin{tabular}{p{0.22\textwidth}p{0.73\textwidth}}
    \toprule
    \textbf{Golden Sentence} & \textbf{Silver Sentences Augmented} \\ 

    \midrule
    \textbf{(Option 3)} I’m a stupid student with no intelligence/future.  &  
    I know I’ll never be like that; I’ll be a stupid failure my entire life.
    \newline
    \newline
    Used to be a stellar student, but I’m scared of opinions now that I received a C in a class.
    \newline
    \newline
    It’s actually starting to irritate me, and I’m starting to feel stupid.
    \\
    \bottomrule
    \end{tabular}}
    \caption{Examples of augmented silver sentences with highest semantic similarity to the golden sentence.}
    \label{table:augmented-sentences-example}
\end{table}


\section{Experimental Settings}
\label{sec:experiments}

We evaluate the performance of our methods in the eRisk2020 and 2021 collections. In eRisk2020, we use 2019 as training data. In eRisk2021, we use the 2019 and 2020 collections as training. The competing methods used the same collection splits, while some of them also considered external datasets (\cref{subsec:experimental-setting}). In our experiments, we study the two components of our approach: $i$) the performance of the semantic retrieval pipeline (\cref{subsec:semantic-retrieval-pipeline}) and $ii$) the effectiveness of the sentence selection strategies (\cref{subsec:silver-labels-generation}). For this reason, our methods consist of combinations of these components.
We consider three hyperparameters: $1)$ The value $k$ of the number of test queries, $Q^{s} = \{q_{1}^{s}, ..., q_{k}^{s}\}$. $2)$ The semantic threshold to select the cut-off of the rankings, $R_{i}^{s}$. $3)$ The number of silver sentences to generate the silver dataset, $Ag^{s}$. The specific hyperparameters and the tuning process are described in Appendix~\ref{appendix-subsec:hyperparameters}.

    \begin{table*}[t]
    \centering
    \scriptsize
    \begin{tabular}{@{}llllclcccccc@{}}
        \toprule
         & &  &  & &\multicolumn{3}{c}{\textbf{Questionnaire Metrics}} & & \multicolumn{2}{c}{\textbf{Symptom Metrics}}\\
        \cmidrule{6-8} \cmidrule{10-11}
        Collection & & \multicolumn{2}{c}{Model} & \makecell{External \\ Dataset} &  DCHR ($\uparrow$) & ADODL ($\uparrow$) & RMSE ($\downarrow$) & & AHR ($\uparrow$) & ACR ($\uparrow$) \\
    
        \midrule
        \multirow{12}{*}{\textbf{eRisk 2020}}

        & &\multicolumn{2}{l}{BioInfo \cite{BIOINFO-2020} (a)}
        & \cmark & 30.00 & 76.01 &  18.78 &  & 38.30 & 69.21 \\  

        & & \multicolumn{2}{l}{ILab \cite{martinez2020early} (b)}
        & \cmark & 27.14 & 81.70  & 14.89 & & 37.07 & 69.41 \\

        & & \multicolumn{2}{l}{Relai \cite{maupome2020early} (c)}
        & \cmark & 34.29 & 83.15  & 14.37 & & 36.39 & 68.32 \\

        & & \multicolumn{2}{l}{UPV \cite{uban2020deep} (d)}
        & \xmark & 35.71 & 80.63 & 15.40 & & 34.56 & 67.44 \\

        & & \multicolumn{2}{l}{Sense2vec \cite{PEREZ2022102380} (e)}
        & \xmark & 37.14 & 82.61 & 12.40 & & \textbf{38.97} & \textbf{70.10} \\        
        
        \cmidrule{3-11}
        & & \textbf{Aggregation} &  \textbf{Silver Sentences Selection} &  &  & &  & &  &   \\
        \cmidrule{3-4}        
        & & Accum Voting & BM25 & \xmark & 38.57$^{a,b,c,d,e}$ & 85.19 & 12.37 & & 35.24 & 68.37  \\
        & & Accum Recall & BM25 & \xmark & 40.00$^{a,b,c,d,e}$ & 84.65 & 12.13 & & 35.71 & 67.60 \\
        
        & & Accum Voting & SBERT & \xmark & 42.86$^{a,c,d}$ & 83.08 & 14.25 & & 34.83 & 65.90 \\
        & & Accum Recall & SBERT & \xmark & 42.86$^{a,b,c,e}$ & 84.51 &  12.37 & & 33.33 & 66.05 \\

        & & Accum Voting & Aug Dataset & \xmark & 47.14$^{a,b,c,d,e}$ &  \textbf{85.33} & \textbf{11.87} & & 35.24 & 67.41 \\
        
        & & Accum Recall & Aug Dataset & \xmark & \textbf{50.00$^{a,b,c,d}$} & 85.24 &  12.09 & & 35.44 & 67.23   \\

        \midrule
        \multirow{10}{*}{\textbf{eRisk 2021}}

        & & \multicolumn{2}{l}{DUTH \cite{spartalis-transfer-2021} (a)}
        & \xmark & 15.00 & 73.97 & 19.60 & & \textbf{35.36} & 67.18 \\

        & & \multicolumn{2}{l}{Symanto \cite{UPV-symanto-2021} (b)}
        & \cmark & 32.50 & 82.42 &  14.46 & & 34.17 & \textbf{73.17}  \\

        & & \multicolumn{2}{l}{CYUT \cite{CYUT-2022} (c)} 
        & \xmark & 41.25 & \textbf{83.59} &  \textbf{12.78} & & 32.62 & 69.46 \\
        
        \cmidrule{3-11}
        & & \textbf{Aggregation} & \textbf{Silver Sentences Selection} &  &   & &  & &  &   \\
        \cmidrule{3-4}          
        
        & & Accum Voting & BM25 & \xmark & 45.00$^{a,b,c}$ & 82.16 &  14.11 & & 30.97 & 64.54 \\
        & & Accum Recall & BM25 & \xmark & 42.50$^{a,b,c}$ & 80.62 & 15.13 & & 28.03 & 62.92 \\

        & & Accum Voting & SBERT & \xmark & 45.00$^{a,b,c}$ & 81.92 & 14.15 & & 29.67 & 64.27 \\
        & & Accum Recall & SBERT & \xmark & 41.25$^{a,b,c}$ & 81.86 & 14.2  & & 27.47 & 62.89 \\

        & & Accum Voting & Aug Dataset & \xmark & 46.25$^{a,b,c}$ & 81.72 & 14.83 & & 27.95 & 62.40  \\
        
        & & Accum Recall & Aug Dataset & \xmark & \textbf{51.25$^{a,b,c}$} & 81.65 & 14.96 & & 27.66 & 61.72 \\
         \bottomrule
    \end{tabular}
    \caption{Results on eRisk collections. The numbers of the official metrics are in percentage. Best values are bolded. Methods using external datasets for training the model are marked. Statistical significant differences in the severity level category assignment according to the Stuart-Maxwell marginal homogeneity test w.r.t to the baselines are super-scripted (p-values < $0.05$). For the remaining metrics, we found no statistically significant differences.}
    \label{table:erisk-results-both-collections}
\end{table*}

\subsection{Datasets}
\label{subsec:erisk-datasets}

The collections selected for experiments correspond to the data delivered for the \textit{eRisk depression severity estimation task} in 2019, 2020 and 2021 editions~\cite{losada2019overview, losada2020erisk, parapar2021overview}. We rely on these collections as they are adopted as the benchmark for this task and contain real answers from Reddit users to the BDI-II. Table~\ref{tab:dataset_stats} summarizes the main statistics of these datasets.

\begin{table}[t]
        \centering
        \small
        \begin{tabular}{lcccc}
        
        \toprule
        \textbf{eRisk Dataset} &  \textbf{Level} & \textbf{2019} & \textbf{2020} & \textbf{2021} \\ 
        \midrule
         \multirow{4}{*}{Users} 
         & \textit{Minimal}   & 4 & 10 & 6        \\
         & \textit{Mild}  & 4 & 23 & 13 \\ 
         & \textit{Moderate}  & 4 & 18 & 27        \\
         & \textit{Severe}  & 8 & 19 & 34        \\
         \cmidrule{3-5}
         Total Users & & 20 & 70 & 80   \\
         
         \midrule
         Avg Posts/User & & 519 & 480 & 404 \\
         Avg Sentences/User & & 1688 & 1339 & 1123 \\
         
        \bottomrule
        \end{tabular}
        \caption{Statistics of the eRisk collections.}
        \label{tab:dataset_stats}
        \vspace{-2mm} 
\end{table}

\subsection{Evaluation}
\label{subsec:experimental-setting}

\textbf{Evaluation Metrics}. We use the official metrics proposed in the eRisk benchmark~\cite{losada2019overview} to keep a fair comparison against the competing methods. These metrics assess the quality of a questionnaire estimated by a system compared to the real one reported by the user. They include two evaluations: $1)$ At questionnaire level, the Depression Category Hit Rate (\textbf{DCHR}) computes the percentage of depressive levels correctly estimated, and the Difference Between Overall Depression Levels (\textbf{ADODL}) computes the overall estimations of the BDI-II score. $2)$ On the other hand, The Average Hit Rate (\textbf{AHR}) and Average Closeness Rate (\textbf{ACR}) assess the results at symptom level. Apart from the eRisk evaluation, we include one additional error metric: the Root-Mean-Square Error (\textbf{RMSE})~\cite{chai2014root} to compare the models predictions of the BDI-II score. Thus, the lower the value reported by RMSE, the lower the difference between predictions and real scores are.

\textbf{Competing Methods.} We consider the best prior works for each metric for the eRisk2020/2021 collections. We refer the reader to the corresponding shared task surveys for a detailed analysis~\cite{losada2020erisk, parapar2021overview}. In eRisk2020, BioInfo~\cite{BIOINFO-2020} and Relai~\cite{maupome2020early} methods obtained their own datasets to perform standard ML classifiers using engineered features as linguistic markers. Other deep learning approaches, such as ILab~\cite{martinez2020early} and UPV~\cite{uban2020deep}, focused their efforts on the use of large language models (LLMs) explicitly trained for depression severity estimation. Finally, a recent work by~\citet{PEREZ2022102380} (Sense2vec) designed different word embedding models for each of the symptoms and achieved state-of-the-art results in this dataset. In eRisk2021, Symanto~\cite{UPV-symanto-2021} team trained a neural model with additional data annotated by psychologists and combined it with a set of engineered features, whereas \citet{CYUT-2022} (CYUT) experimented with different RoBERTa classifiers. Similar to our work, \citet{spartalis-transfer-2021} (DUTH) used semantic features with sentence transformers to extract one dense representation per user, which is then fed as input, experimenting with various classifiers. Although insightful, eRisk approaches cannot evidence the sentences that lead to symptom decisions.

\section{Results and Discussion}

Table \ref{table:erisk-results-both-collections} compares the results of all the variants of our approach against the competing methods. These variants are the combination of our two aggregation methods (\textit{Accum Voting} and \textit{Accum Recall}) and the sentences selection strategies (\textit{BM25}, \textit{SBERT} and the augmented dataset, \textit{Aug Dataset}). The comparison is based on the use of questionnaire and symptom level metrics (\cref{subsec:experimental-setting}).

\textbf{Questionnaire level:} Our approach achieves the best DCHR, which considers the percentage of times that the system estimates the severity level of the users correctly. Most of our variants outperform all prior work in this metric, with the \textit{Accum Recall-Aug Dataset} correctly estimating at least 50\% of the depression levels for both collections. In more detail, it improves $13$ and $10$ points over the best previous results for eRisk2020 and 2021, respectively. A similar phenomenon occurs in the rest of the questionnaire metrics. In the error metric, RMSE, our results also show less estimation error in the BDI-II score.

\textbf{Symptom level}: Although in eRisk2020, our AHR figures are close to the best baselines, that is not the case in 2021. AHR computes the ratio of option responses estimated correctly. The explanation is that we tuned the model hyperparameters for the  DCHR metric since clinicians believe that assessing overall depression levels is more valuable than focusing on specific symptoms~\cite{richter1998validity}. Tuning for AHR may produce worse overall results because the model could be failing to a greater amount in the non-correct answers, resulting in higher overall error. To illustrate that effect, we produced an oracle to obtain the best hyperparameters for each symptom-classifier, maximizing AHR using the \textit{Accum Voting-SBERT} variant. With this oracle, we achieved an AHR of 41.77 and 37.32 for eRisk2020 and 2021, which improves all baselines. However, the oracle obtained worse results in DCHR (24.29 and 36.25). This is because tuning each individual symptom-classifier would require much more training data. We may improve the results for some symptoms with enough data but produce predictions with higher errors (e.g., 0 vs 3) for symptoms with few training samples.

Finally, with respect to the sentence selection strategies, we can observe that using the options descriptions as queries (\textit{BM25} and \textit{SBERT}) performs worse than the augmented dataset (\textit{Aug Dataset}). This emphasizes the importance of a precise candidate selection. Moreover, despite the distribution of depression levels varies in both collections (see Table \ref{tab:dataset_stats}), our methods show robustness as we keep achieving good performance in DCHR.

\label{sec:analysis}

\subsection{Effect of Data Augmentation Strategy}

To better understand the performance of the data augmentation, we report the number of augmented silver sentences along with the F1 metric for each depression level. Table \ref{tab:exp-augmented-sentences-and-f1} shows the F1 results of our best variant using the augmented dataset, \textit{Accum Recall-Aug Dataset}, in eRisk2020 and 2021. Looking at the statistics, we see more presence in golden sentences of high-risk levels (moderate and severe). In addition, the number of silver sentences augmented for each of them is also higher. For example, using eRisk2019 as the training set, an average of three silver sentences were augmented from each golden one in the minimal level ($\frac{310}{98} \approx 3$). In contrast, the average of silver sentences augmented from the severe category is 7 ($\frac{2414}{354} \approx 7$). This suggests that users with higher depressive levels tend to manifest more explicit thoughts related to the symptoms. As a result, our augmentation method finds pieces of evidence in these levels easier.

\begin{table*}[t]
    \centering
    \footnotesize
    \resizebox{\linewidth}{!}{%
    \begin{tabular}{p{0.07\textwidth}p{0.06\textwidth}p{0.06\textwidth}p{0.8\textwidth}}
    \toprule
    \textbf{Symptom} &  \textbf{Golden label} & \textbf{Predicted label} & \textbf{Test queries with more retrieved silver sentences from the predicted label} \\ 
    \midrule
    Sleep problems  & \textbf{1} & \textbf{2} &  
    My sleep cycle consists of staying awake for 48 hours until I can't keep my eyes open.
    \newline
    Same as you, I usually can't go back to sleep once I'm awake. 
    \newline
    I went through a phase where I slept for up to 16 hours (usually partially waking up). \\
    
    \midrule
    Loss of pleasure  & \textbf{3} & \textbf{3} &
    Look, no matter how hard you try, things don't get any better from here.
    \newline
    I don't even enjoy simple things like food that I used to enjoy; there are just foods that I dislike less. 
    \newline
    Why am I not supposed to enjoy life?  \\ 
    
    \bottomrule
    \end{tabular}}
        \vspace{-2mm}
    \caption{Example of the top query sentences from a test user for two symptoms along with the golden option response of that user and the predicted option of our method.}
    \label{tab:test-user-sentences-example}
    \vspace{-2mm}
\end{table*}

\begin{table}[t]
    \centering
    \LARGE
    \resizebox{\linewidth}{!}{%
    \begin{tabular}{p{0.38\textwidth}p{0.75\textwidth}}
    \toprule
    \textbf{Test query} & \textbf{Silver sentences retrieved} \\ 

    \midrule
    My \tcbox{sleep cycle} consists of \tcbox{staying awake} for 48 hours until I can't keep my eyes open.  &  
    \textbf{(Option 2)} Always had \tcbox{trouble sleeping}, no big deal but it's gotten worse in the last two months.
    \newline
    \newline
    \textbf{(Option 2)} I have to get up early to get to university, and I've recently been getting \tcbox{no more than 3-4 hours of sleep.}
    \\
    \midrule
    Look, no matter how hard you try, \tcbox{things dont get any better} from here.&
    \textbf{(Option 3)}  Hoping for a "better thing" \tcbox{never makes me feel better} unless it comes from this sub because I know people get it.
    \newline
    \newline
    \textbf{(Option 3)} \tcbox{Things stop being enjoyable}, and everything becomes a chore. 
    \\
    \bottomrule
    \end{tabular}}
        \vspace{-2mm}
    \caption{Examples of retrieved silver sentences with their assigned label from two test queries from a user.}
    \label{tab:user-case-study-retrieved-sentences}
    \vspace{-2mm}
\end{table}

If we observe the F1 results in Table~\ref{tab:exp-augmented-sentences-and-f1}, we also see considerable variability among depressive levels. In both collections, we achieve better results for higher risk categories. This seems to be related to the number of golden sentences. Therefore, if we obtain more samples belonging to the lower risk levels, there may be an improvement in these categories. Finally, we examine our results with a binary classification setting. For this purpose, we categorize the four depression levels into only two: $1$) \textit{low risk} (minimal + mild levels) and $2$) \textit{high risk} (moderate + severe levels). Table \ref{tab:exp-f1-as-binary-classification} shows the results for the \textit{Accum Recall-Aug Dataset} variant along with the best prior work under this setting. Our results suggest the effectiveness of our method, which distinguishes with fair accuracy between higher and lower risks.

\subsection{Interpretability - Case Study}
\label{sec:case_study}

The lack of reliable clinical markers is one of the barriers to the practical use of mental health prediction models~\cite{10.1093-jamiaopen-ooz054,Amini2020TowardsEI}. By considering a more refined grain in the symptom presence, we provide valuable information that may be strong clinical markers. Table \ref{tab:test-user-sentences-example} showcases how our approach offers interpretability of the symptom decisions, showing three query sentences from an anonymized test user. The symptoms in the Table are \textit{Sleep problems} and \textit{Loss of pleasure}, and the user declared the option $1$ and $3$ for them, respectively. We can see that these test queries are robust indicators of symptom concerns. Following this approach, clinicians may inspect sentences as a first step towards further diagnosis or monitoring methods during treatment.

In addition, Table \ref{tab:user-case-study-retrieved-sentences} displays some of the silver sentences retrieved for the same test queries selected from the anonymized user. The silver sentences are related to the content of the query, and clinicians may evaluate the justifications for every symptom decision by reviewing their labels. Moreover, in our method, false positive/negative predictions can still be helpful for future inspection. For example, for the symptom \textit{Sleep problems}, the test user reported the option $1$, but our method retrieved more silver sentences with the option $2$. While the prediction may be incorrect (golden label $(1)$ $\neq$ predicted label (2)), the risk may still be present.

\section{Conclusions}
\label{sec:conclusions}

We present an effective semantic pipeline to estimate depression severity in individuals from their social media data. We address this challenge as a multi-class classification task, where we distinguish between depression severity levels. The proposed methods base their decisions on the presence of clinical symptoms collected by the BDI-II questionnaire. With this aim, we introduce two data selection strategies to screen out candidate sentences, both unsupervised and semi-supervised. For the latter, we also propose an annotation schema to obtain relevant training samples. Our approaches achieve state-of-the-art performance in two different Reddit benchmark collections in terms of measuring the depression level of individuals. Additionally, we illustrate how our semantic retrieval pipeline provides strong interpretability of the symptom decisions, highlighting the most relevant sentences by semantic similarities.

\section{Ethical Statement}


The collections used in this work are publicly available following the data usage policies. They were collected in a manner that falls under the exempt status outlined in Title 45 CFR §46.104. Exempt research includes research involving the collection or study of existing data, documents, records, or specimens if these sources are publicly available or if the information is recorded by the investigator in such a manner that subjects cannot be identified. We adhered to the corresponding policies and took measures to ensure that personal information could not be identified from the data. The data is available by filling a user agreement according to the eRisk shared task policies\footnote{https://erisk.irlab.org/2021/eRisk2021.html}. In this context, all users have an anonymous state. We paraphrased the reproduced writings to preserve their privacy.

In terms of impact in real-world settings, there is still work to be done to produce effective depression screening tools. The development of such technologies should be approached with caution to ensure that their use is ethical and respects patient privacy and autonomy. Our work aims to supplement the efforts of health professionals rather than replace them. We acknowledge the validation gap between mental health detection models and their clinical applicability. Our goal is to develop automated technologies that can complement current online screening approaches. To ensure safe implementation and as future work, we collaborate with clinicians to validate and obtain a more in-depth analysis of the limitations of these systems.


We took several measures to ensure the objectivity and reliability of our annotations, including providing the same guidelines to all annotators and using a majority vote system to resolve any disagreements. While one of the annotators is also an author of this study, we want to emphasize that they were not given any preferential treatment or guidelines that differed from the other annotators. Moreover, the high agreement percentage among the three annotators (as reported in our study) further supports the reliability and objectivity of our process. Overall, our annotation process was conducted objectively and reliably and the potential for bias was minimized to the greatest extent possible.

Despite being experts in the field, we recognize that annotating depressive symptoms may have an impact on annotators. We provided them with the necessary breaks and did not subject them to any time constraints. Annotators did not report any negative effects after their work. In addition, they were not biased in scoring a higher or lower number of positive sentences.


We recognize that the application of NLP models in real-world scenarios requires careful consideration and analysis due to potential risks and limitations. These models should not be immediately deployed as decision support systems without further studies, including participant recruitment, trials, and regulatory approval. To address potential risks, we consulted with clinical experts to validate the possible dual-use risks before conducting this work and involved them in designing annotation guidelines and all stages of the study. One of the main risks associated with such systems is their performance, as they may produce false positive/negative cases. However, it is important to note that diagnostic discrepancies and their risks associated are common in the clinical setting~\cite{regier2013dsm}.

Therefore, our system is intended to be used in conjunction with health professionals to obtain a more accurate diagnosis. The final decision must always be supported by the validation of a health professional. Our study highlights the potential of NLP-based approaches in assisting clinicians with diagnosis, but further research and testing are needed before it can be considered for clinical deployment.


\section{Limitations}

We recognize that the performance of our solutions is far from ideal to be integrated directly in clinical settings. Moreover, it lacks external validity~\cite{ernala-2019-gap-with-clinical-settings}, as they were never tested in real clinical scenarios. The dataset used in this study (corresponding to the eRisk collections) has a limited size (170 users in total) and diversity, since it only covers one social platform, Reddit. This is partly due to the protection necessary for securing sensitive data related to mental health~\cite{harrigian-etal-2020-models}. We chose the BDI-II questionnaire for our study because it is the only questionnaire with an available dataset that contains both (1) user responses on each symptom and (2) their writing history on social media (Reddit, in this case). Other questionnaires in clinical practice are also widely used (CES-D, GDS, HADS, and PHQ-9). However, we do not have a dataset containing the respondents' answers to the symptoms. As future work, we will work on extending this same pipeline to other related questionnaires.


We are also aware that social media platforms provide an imperfect representation of the population, which is a clear limitation that must be accounted for when using these approaches for public health screening. For this reason, our methods are likely to be modified when other data sources (i.e., other social platforms) are considered. Moreover, when processing data from different clinical contexts (e.g., clinical records), the models may generalize inadequately~\cite{harrigian-etal-2020-models}. Another limitation of our work is the low performance of the symptom evaluation compared with the questionnaire level (related to depression severity levels). As we previously commented, we did not focus our efforts on tuning individual symptom-classifiers but rather to use them as a proxy to estimate depression levels, since we do not have enough training data for most of the symptoms. We also believe that certain errors in these symptom estimations may be due to a lack of awareness of the individual or stigmas associated with the different symptoms.

Despite the gap between mental health prediction models and actual clinical practice, many recent studies~\cite{zhang2022symptom, Zhang-2022-Depression-Scale, yates-etal-2017-depression, PEREZ2022102380} investigated approaches to identifying and detecting depression using reliable clinical markers. We can also see other studies that adhered to clinical questionnaires to investigate other related features such as personality detection~\cite{DBLP:conf/emnlp/YangYQS21}. These studies seek to propose solutions that can be a proxy between health professionals and NLP methods. Our study aims to contribute to this area of research and advance the development of reliable solutions for health professionals.



\bibliography{custom}
\bibliographystyle{acl_natbib}

\appendix

\section{Manual Dataset and Annotation Process}
\label{appendix-sec:annotation-process}

This section describes the construction and annotation schema of our manual dataset. The main idea of this dataset is to obtain a few representative samples that indicate the presence of BDI-II depressive symptoms. For this reason, we develop an annotation schema based on the BDI-II questionnaire~\cite{lasa2000use} to collect a different set of golden sentences belonging to each BDI-II symptom. For each symptom $s$, and the corresponding options $o$, where $o \in \{0, 1, 2, 3\}$, we collect a different set of golden sentences, denoted as $G_{o}^{s}$.

To annotate the golden sentences, we used as data source the training users from the eRisk2019 collection of depression severity~\cite{losada2019overview}. However, the large size of the eRisk collection requires an exhaustive filter for reasonable annotation efforts. For this purpose, we leveraged the data selection strategy of using the option descriptions as queries (\cref{subsec:silver-labels-generation}). In particular, we applied the semantic retrieval variant (SBERT). Using this strategy, we selected candidate sentences for annotating each BDI-II symptom. We have considered this strategy following a recent study that has shown great results in identifying diverse expressions of symptoms for candidate retrieval annotation~\cite{zhang2022symptom}. Previous studies on symptom annotation~\cite{zhang2022symptom} demonstrated a high variance in the distribution of each symptom. For some of them, it is much easier to find representative sentences than for others. To keep the number of annotations per symptom stable, we fixed a similarity threshold of $0.6$ to filter out sentences. However, this similarity threshold still produced too many candidate sentences for some symptoms. For this reason, we further restricted the annotator's work to the first $750$ sentences in the symptoms with too many candidates.

More specifically, 17.15\% of the candidate sentences have been labelled positive following the semantic retrieval strategy from the total of 5004 candidates. From the same labelled sentences, using keyword matching with BM25 reduced this percentage to 4\%. With a random retrieval strategy, it dropped to 0.01\% due to the small number of relevant sentences compared to the size of the entire pool. These findings align with previous research indicating that pattern matching is not effective in retrieving diverse sentences relevant to depressive symptoms~\cite{mowery2017understanding}. Instead, a semantic similarity-based strategy is better suited to retrieve representative sentences without relying on specific keywords covered in the clinical questionnaires.


Following the above candidate annotation schema, we constructed a small dataset for all the BDI-II symptoms. The annotation task was carried out by two psychologists and two PhD students with knowledge in the field. Before the annotation process, we removed all supplementary metadata to avoid bias in the annotators, such as the severity option label ($0-3$) of the user who wrote the sentence. We followed the same annotation procedure as \citet{karisani-2018-health-mentions} to validate the annotation outcomes. This procedure consisted of two phases: $1)$ First, an initial annotator answered the following question in a binary setting (Positive/Negative): \textit{Does the sentence refer to the symptom, and the user talks about himself/herself (first person)?}. This first annotator labelled a total of $738$ positive sentences from the candidate sentences. We considered all the sentences annotated as positive for each symptom to obtain our final labels corresponding to the option levels ($0-3$). Subsequently, we label these positive sentences with the severity option reported by the user who wrote them. Therefore, for each option $o$ and symptom $s$, we obtained a different set of golden labels, $G_{o}^{s}$, where the sentences come from the eRisk users that answered the BDI-II symptoms.

$2)$ Once we had the previous initial annotated sentences, the rest of the annotators validated them. For this purpose, they were provided with a subset containing a random sample of the 20\% of the sentences of each symptom for re-annotation. Since in our pilot experiments, we found much more disagreement with positive labels, the 20\% random sample only contained positive ones. The re-annotation process obtained an 82.44\% among the three annotators, which is an acceptable number considering the sensitivity of this topic \cite{coppersmith-2018}.

Table \ref{table:stats_first_10_symptoms} and \ref{table:stats_last_11_symptoms} show the main statistics of our manual dataset. Visualizing these tables, we can extract several findings. We note that, for all the symptoms, the number of sentences associated with the option $0$ is very low. In some symptoms, even none of the sentences corresponded to option 0. This suggests retrieving sentences representing positive feelings towards the symptom is more complicated. We attribute this fact to two main reasons, ($i$) the descriptions of BDI-II options $0$ are not entirely appropriate for the candidate retrieval process (most of them are just negations of a negative feeling), and ($ii$) users are not as likely to talk about positive as they do with negative feelings. To address this, for the symptoms that lacked sentences with option 0, we manually included between 1 and 3 sentences that provide a positive description of the symptom and labelled them with option 0. 

Finally, the statistics also show that, despite our efforts, there is a clear imbalance in the number of sentences for each symptom and their options. Further details on the dataset will be described with its public release. The dataset will be made available under a research data agreement in accordance with eRisk policies.

\begin{table*}[h]
\scriptsize
\centering
\begin{tabular}{lcccccccccc}
    & \rotatebox[origin=l]{80}{Sadness}
    & \rotatebox[origin=l]{80}{Pessimism}
    & \rotatebox[origin=l]{80}{Sense of Failure}
    & \rotatebox[origin=l]{80}{Loss of Pleasure}
    & \rotatebox[origin=l]{80}{Guilty Feelings}
    & \rotatebox[origin=l]{80}{Punishment}
    & \rotatebox[origin=l]{80}{Self-dislike}
    & \rotatebox[origin=l]{80}{Self-incrimination}
    & \rotatebox[origin=l]{80}{Suicidal Ideas}
    & \rotatebox[origin=l]{80}{Crying} \\

    \midrule
    \textbf{Option 0} & 2 & 2 & 3 & 35 & 2 & 2 & 3 & 1 & 1 & 2\\
    \textbf{Option 1} & 97 & 4 & 6 & 51 & 7 & 0 & 8 & 1 & 29 & 8\\
    \textbf{Option 2} & 8 & 9 & 2 & 18 & 0 & 15 & 42 & 3 & 17 & 32\\
    \textbf{Option 3} & 0 & 44 & 45 & 0 & 23 & 1 & 6 & 5 & 0 & 3\\
    \midrule
    \textbf{Total Labels} & 108 & 59 & 56 & 104 & 32 & 18 & 59 & 10 & 47 & 45\\

    \bottomrule
\end{tabular}
    \caption{Annotations statistics of the first ten BDI-II symptoms.}
    \label{table:stats_first_10_symptoms}
\end{table*}

\begin{table*}[h]
\scriptsize
\centering
\begin{tabular}{lccccccccccc}
    & \rotatebox[origin=l]{80}{Agitation}
    & \rotatebox[origin=l]{80}{Social withdrawal}
    & \rotatebox[origin=l]{80}{Indecision}
    & \rotatebox[origin=l]{80}{Worthlesness}
    & \rotatebox[origin=l]{80}{Loss of energy}
    & \rotatebox[origin=l]{80}{Sleep changes}
    & \rotatebox[origin=l]{80}{Irritability}
    & \rotatebox[origin=l]{80}{Changes in appetite}
    & \rotatebox[origin=l]{80}{Concentration difficulty}
    & \rotatebox[origin=l]{80}{Tiredness/Fatigue} 
    & \rotatebox[origin=l]{80}{Low libido} \\

    \midrule
    \textbf{Option 0} & 2 & 1 & 3 & 1 & 4 & 1 & 2 & 1 & 2 & 3 & 2\\
    \textbf{Option 1} & 7 & 12 & 2 & 1 & 3 & 1 & 26 & 3 & 4 & 15 & 4\\
    \textbf{Option 2} & 4 & 4 & 6 & 3 & 7 & 3 & 4 & 1 & 10 & 4 & 1\\
    \textbf{Option 3} & 13 & 7 & 3 & 19 & 0 & 4 & 0 & 2 & 1 & 3 & 1\\

    \midrule
    \textbf{Total Labels} & 26 & 24 & 14 & 24 & 14 & 9 & 32 & 7 & 17 & 25 & 8\\
    
    \bottomrule
\end{tabular}

    \caption{Annotations statistics of the last eleven BDI-II symptoms.}
    \label{table:stats_last_11_symptoms}
\end{table*}

\section{Detailed Experimental Settings}
\label{appendix-subsec:hyperparameters}

We experimented with different hyperparameters to validate the results from our two main components: the semantic retrieval pipeline (\cref{subsec:semantic-retrieval-pipeline}) and the sentence selection process (\cref{subsec:silver-labels-generation}). As we do not have a validation set, we performed leave-one-out cross-validation using the training set available to calculate the optimal values of all hyperparameters. The metric maximized was \textit{DCHR}. When evaluating our methods in eRisk2020, the training set was the eRisk2019 dataset. When using as test collection the eRisk2021 dataset, the training set was the eRisk2019 and 2020 collections. Table \ref{tab:chosen-hyperparameters-values} presents the hyperparameters and the optimal values for each method used in our experiments~\footnote{We want to note that the tuning of hyperparameters in our method did not result in significant changes to its performance. We thoroughly analysed the impact of hyperparameters on our results and found that the changes were not significant enough to include another section int he article.}.

\textbf{Semantic retrieval pipeline (\cref{subsec:semantic-retrieval-pipeline})}. In the semantic pipeline, we experimented with two hyperparameters:

\textbf{$1)$ The value $k$ of the number of test queries.} We explored with selecting a different number of top $k$ values of the user test queries, $Q^{s} = \{q_{1}^{s}, ..., q_{k}^{s}\}$. To select these test queries, we used the data selection strategies of using the option descriptions as queries (\cref{subsec:silver-labels-generation}). Using BM25, the $k$ values explored were: $[5,10,15,20,25,30,40]$. We also experimented with the same $k$ values using the semantic variant. However, we did not include those results in the paper as they could not improve the use of BM25.

\textbf{$2)$ The semantic threshold to select the cut-off of the rankings,} $R_{i}^{s}$. We experimented with different semantic thresholds to select the cut-off of the ranking of silver sentences, $R_{i}^{s}$.
This semantic threshold, calculated as the cosine similarity, was explored with the next values: $[0.45, 0.50, 0.55, 0.60, 0.65]$. The higher the cosine similarity, the lower the number of silver sentences retrieved by the semantic ranking obtained by the test queries.

\textbf{Silver sentences selection (\cref{subsec:silver-labels-generation})}. Additionally, we also experimented with a filtering hyperparameter for creating more or less restrictive filters when generating the silver dataset, denoted as \textit{selection threshold}. Depending on the selection strategy (BM25, SBERT or Aug Dataset), we used the next sentence selection hyperparameters:

\textbf{$3)$ The number of silver sentences to generate the silver dataset}, $Ag^{s}$. Using BM25, we explored with two different top $k$ values,  $k \in \{50,100\}$ for retrieving the sentences of each training user. In the case of semantic retrieval (SBERT), we explored with the same semantic similarity thresholds as in the semantic ranking:  $[0.45, 0.50, 0.55, 0.60, 0.65]$. Higher cosine similarity implies more restrictions, so the number of silver sentences generated will be lower. Finally, the semantic threshold values explored with the augmented dataset were the same.

With respect to the sentence transformers models \cite{reimers-2019-sentence-bert}, we experimented with different pre-trained models: \textit{msmarco-bert-base-dot-v5\footnote{\url{https://huggingface.co/sentence-transformers/msmarco-bert-base-dot-v5}}, msmarco-distilbert-cos-v5\footnote{\url{https://huggingface.co/sentence-transformers/msmarco-distilbert-cos-v5}}, all-roberta-large-v1\footnote{\url{https://huggingface.co/sentence-transformers/all-roberta-large-v1}} and stsb-roberta-large\footnote{\url{https://huggingface.co/cross-encoder/stsb-roberta-large}}} via the huggingface transformers library. All these models were fine-tuned on diverse semantic similarity datasets. In pilot experiments, the best results were obtained with the model \textit{all-roberta-large-v1}. Thus, all our reported results correspond to the use of that model.

\begin{table*}[h]
    \centering
    \scriptsize 
   \begin{tabular}{ccccc}
        \toprule
         \textbf{Test Set} &\textbf{Method}  & \textbf{Number $k$ of User Test Queries} & \textbf{Semantic Ranking Threshold} & \textbf{Sentence Selection Threshold}  \\
        
        \midrule

        \multirow{4}{*}{\textbf{eRisk 2020}} 
         & Accum Voting - BM25  & 25  & 0.55 & Top k = 100\\
         & Accum Recall - BM25  & 25  & 0.60 & Top k = 100\\
         & Accum Voting - SBERT  & 25  & 0.50 & 0.45\\
         & Accum Recall - SBERT  & 25  & 0.50 & 0.35\\
         & Accum Voting - Aug Dataset  & 40 & 0.55 & 0.50 \\
         & Accum Recall - Aug Dataset  & 35 & 0.55 & 0.50 \\

        \midrule
        
        \multirow{4}{*}{\textbf{eRisk 2021}} 
         & Accum Voting - BM25  & 25 & 0.55 & Top k = 100\\
         & Accum Recall - BM25  & 25 & 0.55 & Top k = 100\\
         & Accum Voting - SBERT  & 25 & 0.50 & 0.50 \\
         & Accum Recall - SBERT   & 30 & 0.55 & 0.40\\
         & Accum Voting - Aug Dataset  & 25 & 0.55 & 0.50\\
         & Accum Recall - Aug Dataset  & 25 & 0.55 & 0.50\\

         \bottomrule
    \end{tabular}
    \caption{Best hyperparameter values for all the variants considered in our methods. These values were obtained by performing leave-one-out cross-validation in the training set by maximizing the DCHR metric.}
    \label{tab:chosen-hyperparameters-values}
    \vspace{-2mm} 
\end{table*}

\end{document}